\documentclass[10pt, a4paper]{article}
\usepackage{lrec2014}
\usepackage{graphicx}
\usepackage{latexsym}
\usepackage{url}
\usepackage{epstopdf}

\title{As Cool as a Cucumber: Towards a Corpus of Contemporary Similes in Serbian}

\name{Nikola Milo\v{s}evi\'{c}, Goran Nenadi\'{c}}

\address{School of Computer Science, University of Manchester \\
               Kilburn building, M13 9PL, Manchester, UK,\\
               nikola.milosevic@manchester.ac.uk, g.nenadic@manchester.ac.uk\\}

\abstract{Similes are natural language expressions used to compare unlikely things, where the comparison is not taken literally. They are often used in everyday communication and are an important part of cultural heritage. Having an up-to-date corpus of similes is challenging, as they are constantly coined and/or adapted to the contemporary times. In this paper we present a methodology for semi-automated collection of similes from the world wide web using text mining techniques. We expanded an existing corpus of traditional similes (containing 333 similes) by collecting 446 additional expressions. We, also, explore how crowdsourcing can be used to extract and curate new similes. \\ \newline }

\begin{document}

\maketitleabstract

\section{Introduction}

Similes are figures of speech which compares two objects through the use of specific connection words \cite{harris2002handbook}, but where comparison is not intended to be taken literally (e.g  "hladan k'o krastavac" -- \textit{as cool as a cucumber}). They are often used as metaphors, but connection words (such as "like" and "as") are used explicitly \cite{cooper1986metaphor,niculae2013comparison}. In many languages, similes are considered as an important building block of spoken language and are part of cultural heritage. In Serbian, for example, Vuk Karad\v{z}i\'c attempted to collect similes, together with other short figures of speech in the XIX century \cite{karadzic1849srpske}. While some of the collected similes are still used (e.g. "bela kao sneg" -- \textit{white as snow}), some have evolved (e.g. "gladan kao kurjak" -- \textit{hungry as a wolf} is now more often appearing as "gladan kao vuk" -- \textit{hungry as a wolf}) and some are rarely used and may not even be understood with the current speakers (e.g. "crveni se kao oderano gove\v{c}e" -- \textit{red as skinned ox}). Of course, a number of new similes have emerged. During the XX century, there were attempts to collect proverbs and other short forms from particular geographical areas \cite{markovic1979metaforisane,jevtic1969narodne,vukanovic1983vranjanske,cvetanovic1980narodna}, but little has been done in updating similes. 

To the best of our knowledge, there is no large-scale text mining approach to collecting similes in Serbian, but there have been attempts to collect them in other languages. For example, Veale and Hao \shortcite{veale2007making} created an approach for automated extraction of explicit similes in English. They modelled them as "X is as P as Y", where X and Y are nouns and P is an adjective.  Bin et al. \shortcite{bin2012using} used similar approach, querying Baidu search engine for Chinese similes. Aasheim \shortcite{Aasheim2012contrastive} differentiates two types of similes: nominal (\textit{like a/an + noun}) and adjectival (\textit{as + adjective + as + noun}). In Slavic languages, simile expressions are based on the preposition "like" or "as" (\textit{kak} in Russian, \textit{kao} in Serbian, \textit{ako} in Slovak, etc.) and introduce a noun phrase (NP) agreeing in case with the standard of comparison \cite{rappaport1998slavic}.

In this paper we explore how we can update the corpus of similes in Serbian using the World Wide Web and crowd-sourcing. The World Wide Web in particular provides people with novel ways to express themselves on a variety of media such as Wikis, question-answer databases, blogs, forums, review-sites and social media. As such, it is an increasingly important resource for exploring linguistic changes and trends, especially as its form is in-between written and spoken languages. Our semi-automated methodology utilises natural language processing, machine learning techniques and manual curation to ensure the validity of extracted data. 

\section{Method}

\subsection{Method Overview}
\begin{figure}[ht!]
\centering
\includegraphics[width=75mm]{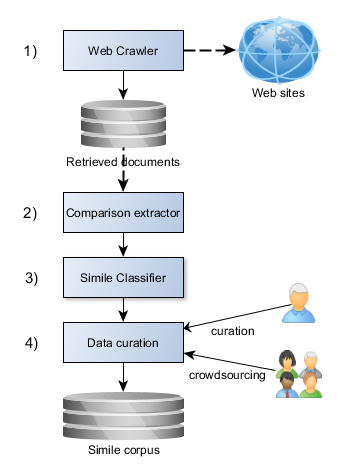}
\caption{Overview of the proposed methodology: 1) Document retrieval using a web crawler, 2) Comparison extraction, based on part-of-speech tagging, 3) Classification of similes; 4) Data curation with a human curator. Users can also propose missing similes.}
\label{Figure 1}
\end{figure}

The methodology consists of four steps (see Figure \ref{Figure 1}): firstly we collect the documents for processing; in the second step we process these documents and extract all comparisons; in the third step, we use machine learning to distinguish similes from the other expressions that function as comparisons; in the final step, a human curator reviews the data and corrects any mistakes. Also, users of our crowdsourcing website can propose new similes and help to collect the similes that were not collected by the text mining pipeline; the curator checks these manually. 

\subsection{Crawler}
In order to collect texts from which to mine similes, we have developed a number of crawlers crafted for particular websites. The aim was to download all potentially meaningful text from these websites, but skip contentless parts or parts where similes are unlikely, such as menus, headings and footers. In order to do that, each crawler extracts only text that is inside the div HTML tag with a particular id. Crawlers follow links on each page they visit, but not outside the domain of the website. 

We have developed separate crawlers for the following four domains:
\begin{itemize}
\item \textit{laguna.rs} - one of the biggest book publishers with abstracts, reviews and parts of the books on their website; 
\item \textit{rastko.rs} - a project that aims to make a digital library of books and articles in Serbian that are considered cultural heritage; 
\item \textit{burek.com} - a large general public forum on Serbian;
\item \textit{tarzanija.com} - a popular blog portal with sarcastic comments on various issues. 
\end{itemize} 

For crawling and processing we used Scrapy, an open source Python framework for extracting data from website\footnote{\url{http://scrapy.org/}}. 

\subsection{Modelling and extracting candidate similes}

The connection word used in similes in Serbian is “kao” - \textit{like, as}”, but it also often appears in shorter forms "k'o" or "ko" (misspelled). There are two main categories of simile in Serbian: 
\begin{itemize}
\item \textbf{Adjectival:} Adjective + Connection word ("kao") + Noun Phrase (e.g. "lep kao cvet" -- \textit{beautiful as a flower})
\item \textbf{Verbal:} Verb + Connection word ("kao") + Noun Phrase (e.g. "radi kao konj" -- \textit{"works like a horse"})
\end{itemize}

Both of these categorise can be modelled by this expression:
\begin{equation} \label{eq:model0}
(V|A|V\ se)\ (kao|ko|k'o)\ (NP)
\end{equation}

A typical noun phrase consists of a noun together with zero or more dependents of various types, such as determiners, adjectives, adjective phrases, noun adjuncts, prepositional phrases, participle phrases or pronouns \cite{crystal2011dictionary}. However, currently publicly available part-of-speech taggers and parsers for Serbian can tag only individual words, but not phrases. Because of this, it was not possible to rely on a tagger to identify noun phrases. We therefore modelled \textit{candidate} similes as a verb or adjective followed by explicit use of connection words "kao", "k'o" or "ko", followed by one or more adjectives and terminated with a noun: 

\begin{equation} \label{eq:model}
(V|A|V\ se)\ (kao|ko|k'o)\ ((A|N)*)\ (N)
\end{equation}

The noun phrase model that we used models the majority of commonly used noun phrases in Serbian. However, noun phrases and similes often contain prepositions. e.g. "smoren kao zmaj u vatrogasnoj stanici" (\textit{bored like a dragon at a fire station}). Our model will currently pick up "smoren kao zmaj" (\textit{bored like a dragon}), which is also a simile used quite frequently, but we would rely on manual curation to extend it with the prepositional phrase.

In the next step of the pipeline, we use retrieved textual documents to extract candidate comparisons. We used a part of speech (POS) model for Croatian and Serbian \cite{agic2013lemmatization} and plugged it into the HunPos tagger \cite{halacsy2007hunpos}. We used the model \ref{eq:model} for matching similes. 

\subsection{Simile classifier}

After a review of initial outputs, we noted a large number of false positives. The model would find similes such as "radi kao konj" (\textit{works like a horse}), but it would also extract expressions like "radi kao pravnik" (\textit{works as a lawyer}), which is not a comparison, but rather specifies someone's profession. There is no lexical feature that would give a cue that these two phrases are in any way different and only semantics (and context) would differentiate the similes from other expressions. However, current publicly available tools for Serbian do not process phrases at the semantic level and thus we decided to rely on a classifier and a final curator's manual review of the phrases.

From the initial data, we have created a sufficient number of true positive and false positive examples that could be used for machine learning. This data was used to create a machine-learning based classifier that can distinguish true similes from the other phrases that have same lexical characteristics. The dataset contained 300 examples of true positive and 300 examples of false positive similes to train and test our algorithm. For the features, we used
\begin{itemize}
\item the whole simile phrase (for illustration we will use "radi kao konj" -- \textit{"works like a horse"}), 
\item the stemmed phrase ("rad ka konj"), 
\item left side of the phrase that is before the connection word ("radi" -- \textit{works}), 
\item left side stemmed ("rad"), 
\item the part of phrase that is on the right from the connection word ("konj" -- \textit{horse}), and 
\item the stemmed right part ("konj"). 
\end{itemize}

For stemming we used stemmer for Serbian \cite{milovsevic2012stemmer}, that was ported for Python\footnote{\url{https://github.com/nikolamilosevic86/SerbianStemmer}}. We experimented with Multinominal Naive Bayes, Random Forests and Support Vector Machines (SVM) with a polynomial kernel that uses sequential minimal optimisation. For machine learning we used Weka toolkit \cite{hall2009weka}.  The models are downloadable from the corpus website\footnote{\url{http://ezbirka.starisloveni.com/download.html}}.

\subsection{Curation interface}

We have created a web portal\footnote{\url{http://ezbirka.starisloveni.com}} for viewing, searching and editing the dataset collected from the previous steps. The portal has the following curation features:

\begin{itemize}
  \item \textbf{Viewing of all currently curated similes}, sorted alphabetically
  
  \item \textbf{Searching for simile.} Searches are performed by using the stemmer for Serbian, so the similes that are not in exactly the same form in the database can be retrieved. Since Serbian is highly inflectional, the aim was to store only one canonical instance (single inflection) of each simile. For example "beo kao sneg" (m), "bela kao sneg" (f) and "belo kao sneg" (n) (\textit{"white as a snow"} in three different grammatical genders - masculine, feminine, neuter), will be treated by our search algorithm as the  expression. Search interface can be seen in Figure \ref{Figure 2}.
  
\begin{figure*}[ht!]
\centering
\includegraphics[width=130mm]{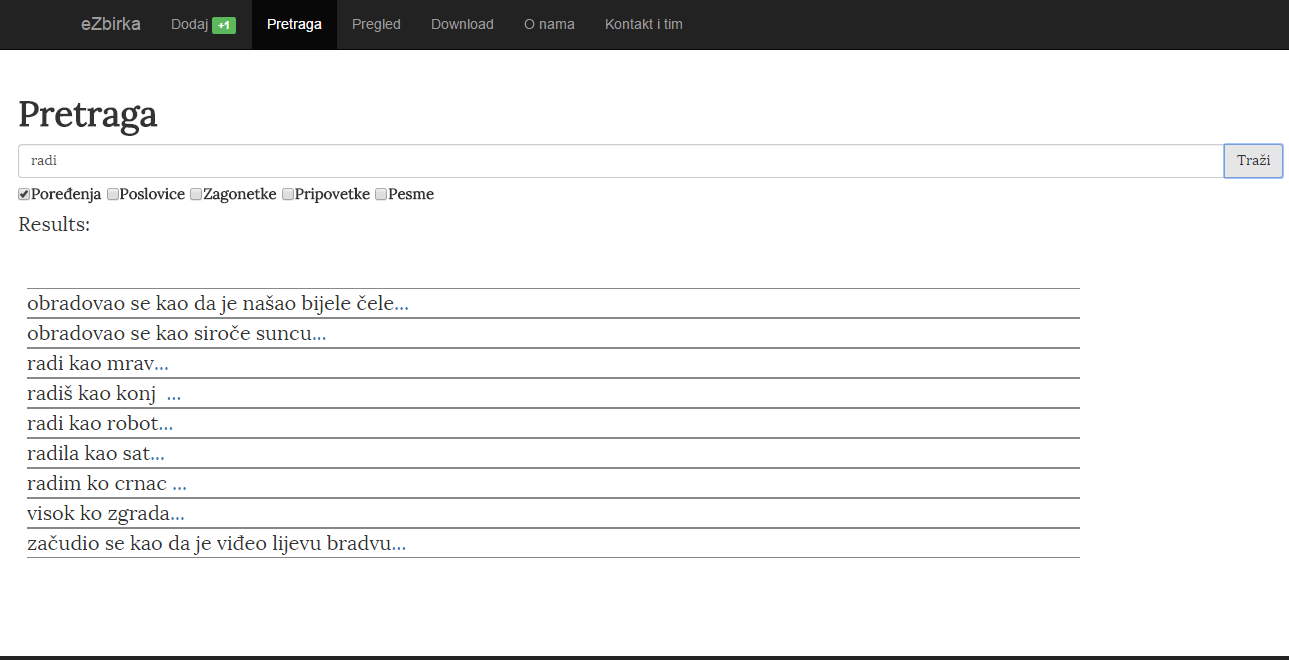}
\caption{Simile search interface}
\label{Figure 2}
\end{figure*}
 
  \item \textbf{Adding a new simile to the database.} Users of the website are able to manually add similes that are rare or missing from the corpus. If the user wants to add simile that already exists in the database, he/she will be notified. Added simile won't be visible on the website until curator approves it.

  \item \textbf{Curation interface.}  Curators can login to the website and perform curation task, such as approving, rejecting or editing similes.
  \end{itemize}

\section{Results}

The crawlers downloaded 40,239 documents from the web (See Table \ref{table:dataset}). Using semi-automated workflow, we extracted potential similes and manually reviewed them. A total of 446 true similes were finally harvested. A manual review of a data also revealed that there were around 5,000 false positives. Our classification approach proved to be useful in filtering some of these out - the results are presented in Table \ref{table:stats}.

\begin{table}[h!]
\centering
\begin{tabular}{ | l  r| }
  \hline
  \small \textbf{Website} & \small \textbf{Number of documents} \\ \hline
  \small burek.com  & \small 22,925 \\ 
  \small laguna.rs & \small 9,574\\ 
  \small tarzanija.com & \small 4,792  \\ 
  \small rastko.rs & \small 2,947  \\ 
  \small Total & \small 40,239  \\ \hline
\end{tabular} 
\caption{Number of documents in dataset obtained by crawlers}
\label{table:dataset}
\end{table}

\begin{table}[h!]
\centering
\begin{tabular}{ | l  r  r  r | }
  \hline
  \small \textbf{Algorithm} & \small \textbf{Precision} &  \small \textbf{Recall} & \small \textbf{F-Measure} \\ \hline
  \small Random Forests  & \small 0.783 & \small 0.732 & \small 0.756  \\
  \small Naive Bayes & \small 0.782 & \small 0.773 & \small 0.777 \\
  \small SVM & \small 0.804 & \small 0.791 & \small 0.797 \\ \hline  
 
\end{tabular} 
\caption{Results of machine learning based classification of similes}
\label{table:stats}
\end{table}

We merged the similes obtained from the web with the simile corpus published by Vuk Karad\v{z}i\'{c} \cite{karadzic1849srpske}, who collected 333 similes. It is interesting that there was only a small overlap between the two simile datasets. Overlap between these two corpora can be seen in Table \ref{table:intersection} The similes collected by  Karad\v{z}i\'{c} were generally more complex, which could be due to the termination rule on noun in our model, so the model might be missing some complex similes. Although, this is an obvious limitation of the system, collection of complex similes will be part of the future work, and also will rely on crowdsourcing interface. At the moment of writing this paper, we had 852 approved similes in our corpus. 

\begin{table}[h!]
\centering
\begin{tabular}{ | l   r | }
  \hline
  \small \textbf{Corpus} & \small \textbf{Number of similes} \\ \hline
  \small WWW Corpus  & \small 446   \\ 
  \small Karad\v{z}i\'{c}'s Corpus & \small 333  \\ 
  \small Intersection between Karad\v{z}i\'{c}'s & \\ \small and WWW Corpus  & \small 9 \\ 
  \small Manually added similes & \small 80 \\ 
  \small Total similes & \small 852 \\ \hline
 
\end{tabular} 
\caption{Number of similes obtained from World Wide Web, Karad\v{z}i\'{c}'s corpus and initial crowdsourcing. Also, number of similes that appear in both WWW and Karad\v{z}i\'{c}'s corpus is presented}
\label{table:corpusstats}
\end{table}

\begin{table}[h!]
\centering
\begin{tabular}{ | l | l  | }
  \hline
  \small \textbf{Original simile} & \textbf{Translation} \\ \hline
	\small beo kao sneg & white as a snow \\
    \small brz kao zec & fast like a rabbit \\
      \small crven kao vampir & red like vampire \\
  \small crven kao krv & red like blood \\
     \small miran kao ovca & calm as a sheep \\
     \small lako kao pero & light as a feather \\
     \small ljut kao ris & angry like a lynx \\
     \small slatko kao med & sweet like honey \\
     \small tvrd kao kamen & heavy like a stone \\ \hline
 
\end{tabular} 
\caption{Intersection between similes found on the WWW and in  Karad\v{z}i\'{c}'s collection}
\label{table:intersection}
\end{table}

\section{Summary}

In this paper we present initial work towards building a corpus of similes in Serbian and a methodology for a semi-automated creation of simile corpus. We have kick-started the process by processing a set of web sites that are likely to have a number of similes. The current corpus, with 852 similes is, to the best of our knowledge, the largest simile corpus in Serbian. We believe that, by mining more web pages for similes and by the use of crowdsourcing, the corpus will grow and remain up to date. Still, the current methodology will need to be improved through the use of semantic resources (e.g. WordNet) and contextualisation.

The International Standard Language Resource Number (ISLRN) of the Serbian Simile Corpus is \textit{534-837-568-404-3}. It is freely available at 
\textit{\url{http://www.ezbirka.starisloveni.com/}}. The code of the simile extractor can be found on GitHub\footnote{\url{https://github.com/nikolamilosevic86/SerbianComparisonExtractor}}.

\bibliographystyle{lrec2014}
\bibliography{similes}

\end{document}